\documentclass[letterpaper, 10 pt, journal, twoside]{IEEEtran}
\usepackage{amsmath,amsfonts}
\usepackage{algorithmic}
\usepackage{array}
\usepackage{subfig}
\usepackage{textcomp}
\usepackage{stfloats}
\usepackage{url}
\usepackage{verbatim}
\usepackage{graphicx}

\usepackage{cite}
\usepackage{booktabs}
\usepackage{multicol}
\usepackage{makecell}
\usepackage{multicol}
\usepackage{amssymb}

\usepackage{tikz}
 
\def\BibTeX{{\rm B\kern-.05em{\sc i\kern-.025em b}\kern-.08em
    T\kern-.1667em\lower.7ex\hbox{E}\kern-.125emX}}
\usepackage{balance}

\newcommand\copyrighttext{%
   \textcopyright 2022 IEEE. Personal use of this material is permitted. Permission from IEEE must be
obtained for all other uses, in any current or future media, including
reprinting/republishing this material for advertising or promotional purposes, creating new
collective works, for resale or redistribution to servers or lists, or reuse of any copyrighted
component of this work in other works. 
  }
\newcommand\copyrightnotice{%
\begin{tikzpicture}[remember picture,overlay]
\node[anchor=south, font=\fontsize{6}{10}\selectfont, yshift=10pt] at (current page.south) {\fbox{\parbox{\dimexpr\textwidth-\fboxsep-\fboxrule\relax}{\copyrighttext}}};
\end{tikzpicture}%
}

\begin{document}

\title{Visual-Inertial Odometry with Online Calibration of Velocity-Control Based Kinematic Motion Models}
\author{Haolong Li$^{1}$, Joerg Stueckler$^{1}$%
	\thanks{Manuscript received: November, 12, 2021; Revised February, 19, 2022; Accepted March, 29, 2022.}
	\thanks{This paper was recommended for publication by Editor E. Marchand upon evaluation of the Associate Editor and Reviewers' comments.
		This work was supported by Max Planck Society and the Cyber Valley Research Fund (CyVy-RF-2019-05). The authors thank the International Max Planck Research School for Intelligent Systems (IMPRS-IS) for supporting Haolong Li.
		We thank Felix Grueninger from Robotics ZWE at MPI-IS for building the robot used in our experiments.} 
	\thanks{$^{1}$All authors are with the Embodied Vision Group,
		Max Planck Institute for Intelligent Systems, T\"ubingen, Germany
		{\tt\footnotesize \{haolong.li,joerg.stueckler\}@tue.mpg.de}}
	\thanks{Digital Object Identifier (DOI): see top of this page.}
}

\markboth{IEEE Robotics and Automation Letters. Preprint Version. Accepted March~2022}%
{Li \MakeLowercase{\textit{et al.}}: Visual-Inertial Odometry with Online Calibration of Velocity-Control Based Kinematic Motion Models}

\maketitle

\begin{tikzpicture}[remember picture, overlay]
    \node [align=center, font=\fontsize{6}{10}\selectfont, yshift=-0.5cm] at (current page.north) {This article has been accepted for publication in IEEE Robotics and Automation Letters. This is the accepted version which has not been fully edited and \\
content may change prior to final publication. Citation information: DOI 10.1109/LRA.2022.3169837};
\end{tikzpicture}

\copyrightnotice

\begin{abstract}
	
	Visual-inertial odometry (VIO) is an important technology for autonomous robots with power and payload constraints. 
	In this paper, we propose a novel approach for VIO with stereo cameras which integrates and calibrates the velocity-control based kinematic motion model of wheeled mobile robots online. 
	Including such a motion model can help to improve the accuracy of VIO.
	Compared to several previous approaches proposed to integrate wheel odometer measurements for this purpose, our method does not require wheel encoders and can be applied when the robot motion can be modeled with velocity-control based kinematic motion model.
	We use radial basis function (RBF) kernels to compensate for the time delay and deviations between control commands and actual robot motion. The motion model is calibrated online by the VIO system and can be used as a forward model for motion control and planning.
	We evaluate our approach with data obtained in variously sized indoor environments, demonstrate improvements over a pure VIO method, and evaluate the prediction accuracy of the online calibrated model.
	
\end{abstract}


\section{Introduction}
\IEEEPARstart{I}{n} recent years, visual-inertial odometry has seen tremendous progress (e.g.~\cite{mourikis2007_msckf,leutenegger2015_okvis,usenko19nfr,campos2021_orbslam3}), driven by the many potential applications of such technology for augmented/virtual reality and autonomous robots, in particular flying robots.
Surprisingly, for wheeled robots, VIO is not trivial to employ due to observability limitations for planar linear motions~\cite{wu2017_vinsonwheels}.
This can be alleviated by integrating motion model constraints into the state estimate. 
A popular approach in the literature is to use wheel odometer measurements to this end (see e.g.~\cite{wu2017_vinsonwheels,ma2019_ack-msckf,yang2019_observability}).

In this paper, we take a conceptually different approach by integrating a velocity-control based kinematic motion model which does not rely on wheel encoders.
By integrating the model into the state estimate, the model parameters such as the relative position of the sensor on the robot or the offset between model and real robot can be calibrated online.
Differently to wheel odometry based models, velocity-control based models can be directly used for downstream tasks such as model-predictive motion control and planning~\cite{Kavraki96},~\cite{LaValle99},~\cite{Kuwata08motionplanning}.
We base our method on a non-linear optimization based approach~\cite{usenko19nfr} to visual-inertial odometry which optimizes state variables such as sensor pose and velocity, IMU biases, and keypoint map in a window of recent frames.
Old frames and IMU measurements are marginalized in a proper probabilistic way to maintain the prior observations as prior knowledge.
The IMU measurements are preintegrated into relative motion measurements between frames.
In this framework, we include a velocity based motion model which models the motion of a wheeled robot in a plane based on linear forward and rotational velocity controls.
For accurate integration of the measurements and controls, an accurate calibration of the sensor placement with respect to the drive, the time synchronization of the controls relative to the visual and inertial measurements, and an identification of the effect of control commands for the underlying low-level robot motion controller on actual executed motion are required.
To model the unknown properties of the controller, we aggregate the raw control commands with a kernel function.
We add parameters for the kernels and the placement (extrinsics) of the sensor on the robot to the estimation problem.
The parameters are calibrated online in the non-linear optimization framework. 

We evaluate our approach on data obtained with a mobile robot in several sizes of environments.
We demonstrate that incorporating the velocity-control based kinematic motion model improves the accuracy and robustness of the VIO estimate.
Moreover, we provide results on the prediction accuracy of our online calibrated model for reference for model-based control and planning approaches.

In summary, our contributions are:
\begin{itemize}
	\item We propose a novel visual-inertial odometry approach for wheeled robots which includes a velocity-control based kinematic motion model into the state estimate.
	\item The parameters of the motion model are calibrated online with the VIO estimate.
	\item We demonstrate that inclusion of the motion can improve the VIO estimate in various indoor environments of different sizes. We also provide evaluation of the prediction accuracy of the calibrated model.
\end{itemize}
Our model can be an alternative to wheel-odometry based methods when a velocity-control based model should be directly calibrated for use in model-predictive control and motion planning methods.

\section{RELATED WORK}

Motion estimation by fusing odometry, IMU and camera data has recently spurred significant interest by the robotics community due to its applications for inertial navigation systems in service robotics and autonomous driving.

\subsection{Inertial and inertial-wheel odometry}

Various recent approaches combine IMU and wheel odometry. Brossard et al.~\cite{brossard2020_ai-imu} suggest an EKF based approach which uses deep learning to predict the noise properties in an EKF framework which fuses IMU measurements to predict motion.
In~\cite{brossard2019_learning}, deep kernel Gaussian Process models are learned for the motion and observation models which are used to fuse IMU and wheel odometry measurements in an EKF framework.
In RINS-W~\cite{brossard2019_rins-w}, Brossard et al. propose to estimate the motion from IMU and odometry measurements using an RNN which detects different motion profiles in an EKF framework.
These approaches, however, do not use the complementary strengths of visual measurements and do not provide a forward model.

\subsection{Visual-inertial-wheel odometry}

Cameras provide complementary information to inertial measurement units for motion estimation.
For general motions, 3-DoF linear acceleration and rotational velocity measurements make roll and pitch orientation observable relative to the gravity direction.
However, double integration of the linear acceleration requires accurate estimation of biases (offsets) in the measurements and makes linear position estimation prone to drift.
Similarly, the yaw orientation around the gravity direction is not observable and prone to drift due to noisy and biased gyroscope measurements.
Visual measurements provide a reference for pose estimation to a local 3D map of the environment which is concurrently build with the pose estimates in VIO approaches.
By this, all DoFs become observable, while the IMU provides high frame-rate measurements which improve the accuracy between images.

The Multi-State-Constrained Kalman Filter (MSCKF~\cite{mourikis2007_msckf}) for VIO has been recently extended to incorporate wheel odometry and overcome observability issues of monocular visual-inertial odometry on wheeled robots in VINS on wheels~\cite{wu2017_vinsonwheels}.
The authors analyze observability of monocular visual-inertial navigation systems on a mobile robot platform and show that for specific motions, scale and 3-DoF rotation become unobservable. 
They also show that adding wheel encoder measurements makes scale observable. 
The approach does not calibrate the motion model parameters online like our method.
In our setting, scale is already observable through the fixed calibrated baseline of our stereo camera. 

Ma et al.~\cite{ma2019_ack-msckf} adopt the VINS on wheels approach and extend it with an Ackerman drive model. Jung et al.~\cite{jung2020_monocular} add GPS measurements directly to VINS on wheels to make position observable.
Yang and Huang~\cite{yang2019_observability} analyze observability for VINS on wheels with line and plane observations.
Another approach concurrently estimates the wheel slippage with VIO~\cite{dang2018_wheelslip}.
More closely related to our method is the approach by Lee et al.~\cite{lee2020_viwo_onlinecalib} which investigates online calibration of the wheel odometry parameters and analyzes observability of the calibration parameters for different constraint motion scenarios.
In contrast, we employ a non-linear optimization based approach for visual-inertial odometry and incorporate an inverse motion model for constraints.

Some approaches integrate wheel odometry into non-linear optimization based approaches.
Liu et al.~\cite{liu2019_viwo} develop online calibration of the extrinsics between camera, IMU and odometer.
Liu et at.~\cite{liu2021_bidirectional_viwoinit} propose a novel initialization approach which corrects the initial state estimates after the first turning motion to handle unobservability of the calibration parameters for straight motions.
Chen et al.~\cite{chen2019_perception} calibrate visual-inertial-wheel odometry offline.
The approach in~\cite{zhang2018_comparison} integrates smooth motion manifold constraints.
Zheng and Liu~\cite{zheng2019_se2xyz} incorporate a planar motion constraint for the mobile base but allow small deviations from this motion in 6 DoF for the camera-IMU system. 
Also, different to ours, these approaches use wheel encoders to measure odometry.
We propose a new approach which incorporates an inverse motion model into optimization-based visual-inertial odometry and calibrates the model online including extrinsics and time synchronization.

\subsection{Learning dynamics models for control}

Optimal control approaches typically rely on action-conditional dynamics models which use them as forward models to plan towards goals.
One recent example is the approach of Williams et al.~\cite{williams2016_aggressivedriving} which learns a dynamics model offline from GPS.
In~\cite{kabzan2019_mpc_racing} a sensor-based localization method using LiDAR, optical speed sensors and INS is used to provide feedback for learning deviations from an analytic dynamics model.

In our approach, we tightly integrate a kinematic motion model in a visual-inertial odometry which allows for calibrating the model online.

\section{METHOD}

We integrate a velocity-control based mobile robot motion model into visual-inertial odometry and optimize the parameters of the motion model concurrently with the camera trajectory and bias parameters of the IMU.
The motion model further constrains the camera motion estimate.
The calibrated motion model could be useful for model-predictive motion control and path planning.

\begin{figure}[tb!]
	\centering
	\includegraphics[width=0.99\linewidth]{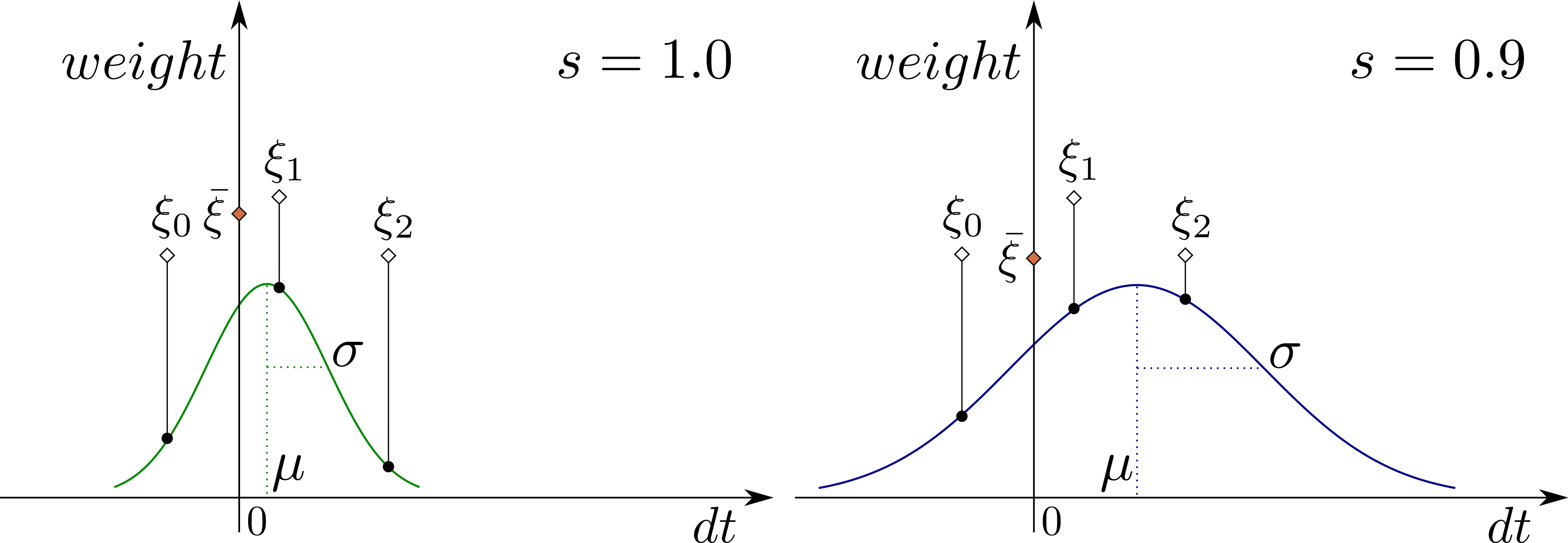}\\\vspace{1ex}
	\includegraphics[width=0.99\linewidth]{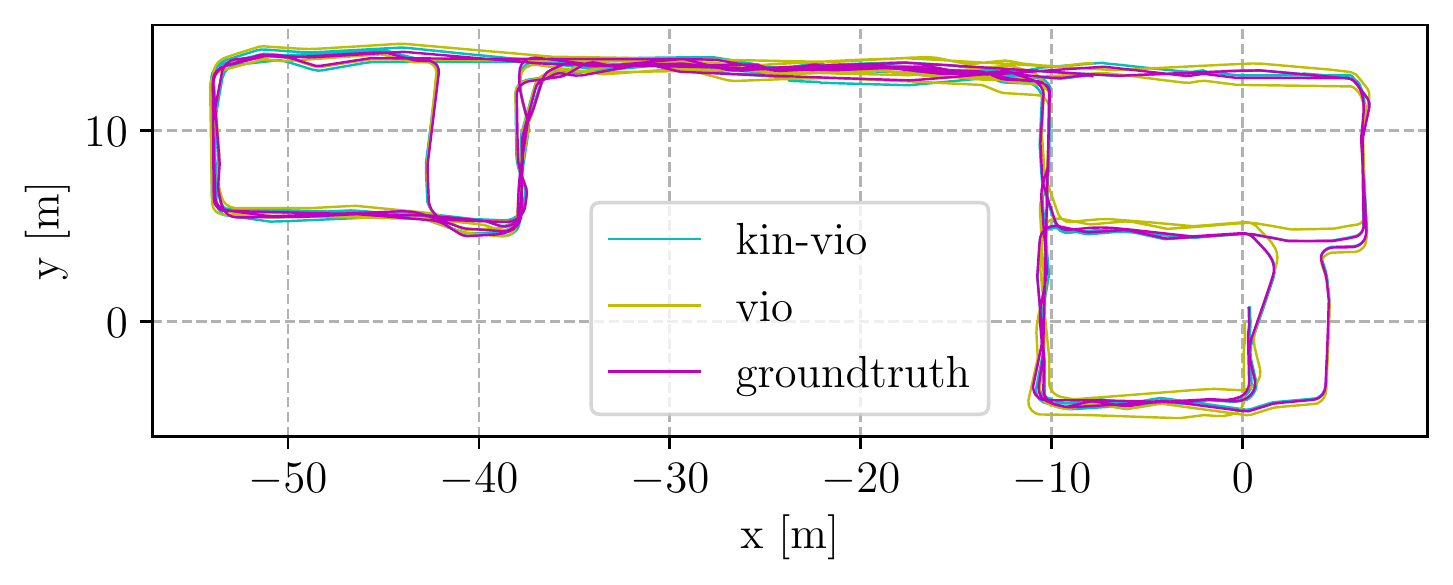}	
	\caption{Weighted aggregation of effective controls and an exemplary result of motion-constrained VIO (\emph{large-01}). Top: Time delays of controls and hardware restrictions (such as acceleration limits) can be handled implicitly by weighting and averaging the commands in a window with an RBF kernel. We optimize for the mean value $\mu$, the variance $\sigma$ and the scale $s$ to shift the kernel and change its shape. Bottom: Our motion-constrained VIO approach achieves smaller deviation with respect to the ground truth. The pure VIO result is shown in yellow, our kinematics-constraint VIO estimate in cyan, ground truth in purple. }
	\label{fig:rbf_traj}
\end{figure}

\begin{figure}[tb]
	\centering
	\includegraphics[width=0.70\linewidth]{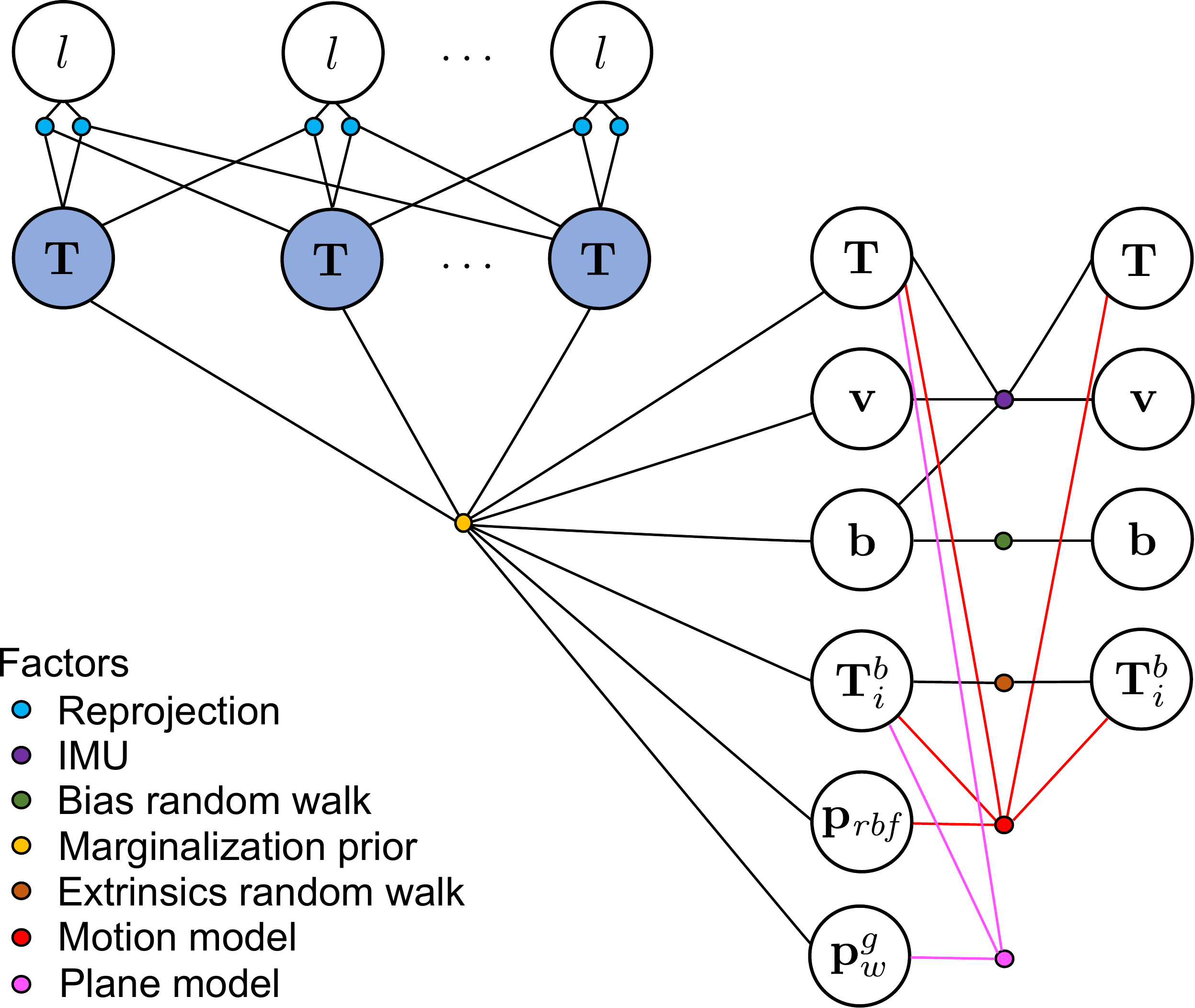}	
	\caption{Factor graph of the proposed method, where $\mathbf{T}^b_i$ is the extrinsic pose. The motion factor consists of frame poses, extrinsic poses and RBF parameters $\mathbf{p}_{\mathit{rbf}}$, and the plane factor includes the frame poses, extrinsic poses and plane parameters $\mathbf{p}_w^g$.}
	\label{fig:graph}
\end{figure}

\subsection{Visual-Inertial Odometry}

We extend the non-linear optimization-based visual-inertial odometry approach in~\cite{usenko19nfr}.
The approach uses a KLT-based keypoint tracking frontend to track the camera motion from frame to frame.
Keyframes are extracted along the camera trajectory and the keypoint tracks generate landmarks in the keyframes with corresponding point measurements. 
Optimization variables are the camera poses for the frames and keyframes, and the camera velocity and biases in the frames.
Only a set of recent keyframes and frames is optimized (3 frames and 7 keyframes in our experiments).
Older frames and keyframes are marginalized from the optimization window and their information serves as a prior.
More formally, for each frame at time $t$, we estimate the camera pose $\mathbf{T}_t \in SE(3)$, the camera velocity $\mathbf{v}_t$, and the bias parameters $\mathbf{b}_t$ of the IMU.
For the keyframes, their camera poses $\mathbf{T}$ are optimized.
The tracked keypoints become landmarks which are hosted in the keyframe in which they have been observed for the first time.
Landmark~$l$ is parameterized by 2D coordinates $(u_l,v_l)$ which minimally represent the bearing vector to the 3D point, and the inverse distance $d_l$. 
The KLT tracking provides measurements of the landmarks in subsequent frames and keyframes.
Residuals
$\mathbf{r}_{tl} = \mathbf{z}_{tl} - \pi\left( \mathbf{T}_t, \mathbf{T}_{h(l)}, u_l, v_l, d_l \right)$
are determined which measure the difference between the measured 2D positions $\mathbf{z}_{tl}$ in image at time $t$ towards the reprojection of landmark~$l$ hosted in frame $h(l)$ into the image using function $\pi$.

In addition to visual residuals, an IMU provides linear acceleration and rotational velocity measurements from which further residuals on the pose and velocity estimates are formed.
To this end, the IMU measurements are preintegrated~\cite{ForsterCDS15} to obtain pseudo-measurements $\boldsymbol{\Delta} \mathbf{T}$, $\boldsymbol{\Delta} \mathbf{v}$ between successive frames at times $t_i$, $t_{j}$ with associated uncertainty $\mathbf{\Sigma}_{ij}$.
The inertial residuals are
\begin{align}
\mathbf{r}_{ij,\boldsymbol{\Delta} \mathbf{R}} &= \log\left( \boldsymbol{\Delta} \mathbf{R} \mathbf{R}_j^\top \mathbf{R}_i \right),\\
\mathbf{r}_{ij,\boldsymbol{\Delta} \mathbf{p}} &= \mathbf{R}_i^\top (\mathbf{p}_j - \mathbf{p}_i - \frac{1}{2} \mathbf{g} \Delta t^2) - \boldsymbol{\Delta} \mathbf{p},\\
\mathbf{r}_{ij,\boldsymbol{\Delta} \mathbf{v}} &= \mathbf{R}_i^\top (\mathbf{v}_j - \mathbf{v}_i - \mathbf{g} \Delta t) - \boldsymbol{\Delta} \mathbf{v},
\end{align}
where $\mathbf{g}$ is the gravity direction and $\Delta t = t_j - t_i$.

The visual-inertial odometry corresponds to the non-linear optimization problem
\begin{equation}
\text{min}~E_{\mathit{VIO}} = \sum_{t \in \mathcal{K} \cup \mathcal{F}} \sum_{l \in \mathcal{O}(t)} \mathbf{r}_{tl}^\top \mathbf{\Sigma}_{tl}^{-1} \mathbf{r}_{tl} + \sum_{(i,j) \in \mathcal{C}} \mathbf{r}_{ij}^\top \mathbf{\Sigma}_{ij}^{-1} \mathbf{r}_{ij}
\end{equation}
where $\mathcal{K}$ and $\mathcal{F}$ are the set of keyframes and frames, respectively, $\mathcal{O}(t)$ is the set of landmarks observed in frame $t$, and $\mathcal{C}$ is the set of indices of subsequent frame pairs.
To optimize the error function $E$ efficiently, old keyframes and frames outside an optimization window are marginalized and only the variables inside the window are optimized.

For further details on the visual-inertial odometry method, we refer the reader to~\cite{usenko19nfr}.

\subsection{Velocity-Control Based Motion Model} 
We use a velocity-control based motion model~\cite{thrun2005probabilistic} and assume that the robot can be controlled by a control command $\mathbf{u} = ( v, \omega )^\top$ through a linear velocity $v \in \mathbb{R}$ in forward direction and a rotational velocity $\omega \in \mathbb{R}$.
The motion model propagates the robot pose $\mathbf{P}_{t} \in SE(2)$ on the ground plane with the control command,
$\mathbf{P}_{t'} =  \mathbf{P}_{t} \exp\left( \Delta t \, \widehat{\boldsymbol{\xi}}  \right)$,
where $\boldsymbol{\xi} = ( v, 0, \omega )$ is the twist vector, $\widehat{\boldsymbol{\xi}} = \left( \begin{array}{ccc} 0 & -\omega & v\\ \omega & 0 & 0\\ 0& 0& 0 \end{array} \right)$ maps a 3D vector to $se(2)$, and 
\begin{multline}
\exp\left( \Delta t \, \widehat{\boldsymbol{\xi}}  \right)
=  \mathbf{P}_{t'}^{t} = \\
\left( \begin{array}{ccc}
\cos(\omega \Delta t) & -\sin(\omega \Delta t) & \frac{v}{w} \sin(\omega \Delta t)\\
\sin(\omega \Delta t) & \cos(\omega \Delta t) & \frac{v}{w} - \frac{v}{w} \cos(\omega \Delta t)\\
0 & 0 & 1
\end{array} \right),
\end{multline}
is the exponential map of $SE(2)$ with time difference $\Delta t$ between the successive time steps.
The exponential map finds the relative $SE(2)$ motion for constant velocities $\boldsymbol{\xi}$ over time duration $\Delta t$.
The logarithm map $\widehat{\boldsymbol{\xi}} = \log( \mathbf{P} )$ of $SE(2)$
is the inverse of the exponential map and maps relative poses $\mathbf{P} \in SE(2)$ to Lie algebra elements in $\widehat{\boldsymbol{\xi}} \in se(2)$. 
The operator $\widehat{\cdot}$ maps 3D twist coordinate vectors to twists in $se(2)$.

\subsection{Motion Model Residuals}

We incorporate the velocity-control motion model into the visual-inertial odometry framework in order to calibrate the parameters of the model and improve the robustness of the VIO.
There are basically two choices to form residuals with a motion model: relative pose constraint (forward) or velocity constraint (inverse).
Mathematically the forward and inverse model residuals are equivalent to each other.
Both kinds of residuals achieve similar results, while the Jacobian matrix of the forward model is computationally more expensive (see~appendix).
For our motion constraint, we treat the velocity controls as measurement and assume the measurement noise comes from the velocity controls. 
In this case, using the inverse model residuals is simpler while using the forward model residuals requires error propagation with linear approximation of the exponential mapping.

Velocity commands are executed by the robot in the robot base frame whose pose relative to the world frame is denoted by $\mathbf{T}_b^w \in SE(3)$ (transforming coordinates from base $b$ to world frame $w$).
In the base frame, the $x$-axis points in forward driving direction, while the $z$-axis points upwards and is the axis of rotational robot motion.
The VIO provides pose estimates of the body frame of the IMU-camera sensor in the world frame, i.e. $\mathbf{T}_i^w \in SE(3)$.
The sensor is placed rigidly on the robot at a relative pose $\mathbf{T}_i^b \in SE(3)$ to the robot base frame.
To quantify the relative motion $\mathbf{T}_{b,t'}^{b,t}$ of the robot base frame from times $t$ to $t'$ of subsequent image frames, we can hence determine
$\mathbf{T}_{b,t'}^{b,t} =  \mathbf{T}_{i,t}^{b,t} \left(\mathbf{T}_{i,t}^{w}\right)^{-1} {\mathbf{T}_{i,t'}^{w}} \left(\mathbf{T}_{i,t'}^{b,t'}\right)^{-1}$.
The rotation $\Delta \theta$ around the $z$-axis of the base frame is calculated from the relative rotation $\mathbf{R}_{b,t'}^{b,t}$ in $\mathbf{T}_{b,t'}^{b,t}$ as the $z$-component of $\log\left( \mathbf{R}_{b,t'}^{b,t} \right)$.
The translational motion $( \Delta x, \Delta y )^\top$ in the $x$-$y$-plane is determined from the corresponding entries of $\mathbf{T}_{b,t'}^{b,t}$. 
The estimated twist is 
\begin{equation}
\boldsymbol{\zeta} = \\ \frac{1}{\Delta t} \log  \left( \begin{array}{ccc}
\cos(\Delta \theta) & -\sin(\Delta \theta) & \Delta x\\
\sin(\Delta \theta) & \cos(\Delta \theta) & \Delta y\\
0 & 0 & 1
\end{array} \right),
\end{equation}
where $\Delta t = t' - t$.
We add residuals of the form
$\mathbf{r}_{\boldsymbol{\xi}} = \boldsymbol{\zeta} - \boldsymbol{\bar{\xi}}$
which implicitly measure the difference between the state estimates and the motion model prediction.

\subsection{Effective Control Command}
In practice, the real action of the robots differs from the received control commands due to effects such as time offsets and properties of low-level controllers. 

Typically, control inputs and image frames are not synchronized but run asynchronously and often also at different rates. 
In our experiments, the control rate is 15\,Hz and is lower than the 30\,Hz image frame rate which is also used to update the VIO estimate.
Moreover in the real world, a delay exists between the control command sent by the controller and the control command executed by the robot. 

The robot physical hardware acceleration limits and internal controllers also prevent the robot from directly executing the control command even if the delay is known. 
To mitigate this difference and build a meaningful residual, we estimate an effective control $\boldsymbol{\bar{\xi}}_t$ at arbitrary time $t$, e.g. at the time of an image frame, from a window of most recent commands.
We average a window of recent control commands with weights determined by a RBF kernel (see Fig.~\ref{fig:rbf_traj}) for the translational and rotational parts separately:
\begin{equation}
\boldsymbol{\bar{\xi}}_t = \\ \left(
\begin{array}{c}
s_{lin} \frac{\sum_{\tau \in \mathcal{W}_t} \exp\left( - \frac{\left\| d\tau - \mu_{lin} \right\|^2}{2 \sigma_{lin} ^2} \right) v_{\tau}} {\sum_{\tau \in \mathcal{W}_t} \exp\left( - \frac{\left\| d\tau - \mu_{lin} \right\|^2}{2 \sigma_{lin} ^2} \right)}\\
0 \\
s_{ang} \frac {\sum_{\tau \in \mathcal{W}_t} \exp\left( - \frac{\left\| d\tau - \mu_{ang} \right\|^2}{2 \sigma_{ang} ^2} \right) w_{\tau}} {\sum_{\tau \in \mathcal{W}_t} \exp\left( - \frac{\left\| d\tau - \mu_{ang} \right\|^2}{2 \sigma_{ang} ^2} \right)}
\end{array} \right) .
\end{equation}
Here $\mathcal{W}_t$ is a window of $N$ control commands indexed by their times $\tau$ at or before time $t$ and $d\tau := t - \tau$. 
For an image frame at time $t$, the window typically spans the $N$ control commands that have occurred before the frame.
We optimize for $\mu$ and $\sigma$ and scale factor $s$ of both linear and angular parts as global parameters together with the VIO states. 
The RBF parameters are summarized in the state variables $\mathbf{p}_{\mathit{rbf},t}$ at time $t$.
In the experiments we demonstrate that the optimized RBF kernel can be used for motion prediction.

\subsection{Motion-Model-Based Error Function Terms}
The robot body can vibrate during operation, the extrinsic pose $\mathbf{T}^b_i$ is thus modeled as a time-variant state that is affected by white noise. 
For $k$ factors within the optimization window this kinematics-based error can be summarized as
\begin{equation}
E_{\mathit{kin}} = \sum_{k} \mathbf{r}_{\boldsymbol{\xi},k}^\top \mathbf{\Sigma}_{\boldsymbol{\xi},k}^{-1} \mathbf{r}_{\boldsymbol{\xi},k} + 	\sum_{k} \mathbf{r}_{\mathit{extr},k}^\top \mathbf{\Sigma}_{\mathit{extr},k}^{-1} \mathbf{r}_{\mathit{extr},k},
\end{equation}
where $\mathbf{\Sigma}_{\boldsymbol{\xi},k}$ is the diagonal weight matrix for the velocity residuals, $\mathbf{r}_{\mathit{extr},k}$ is the difference between two adjacent extrinsic pose estimates and $\mathbf{\Sigma}_{\mathit{extr},k}$ is the diagonal weight matrix that reflects the white noise.

The VIO system takes the image frame at time $t'$ and the raw controls up to the time $t$ of the previous frame as input for the optimization which also calibrates the RBF parameters for the motion model. 
The controls can be generated by manual control or an automatic high-level controller such as model predictive control for path tracking.
In our experiments we use manual control commands as input and calibrate the RBF parameters online. 
A high-level controller would potentially require extrapolation of the last state estimate at the image rate to the current control time using the previous controls and the motion model.

\subsection{Plane Motion Constraint}
We exploit prior knowledge that our robot moves on flat ground in indoor environments and add a stochastic
plane constraint~\cite{wu2017_vinsonwheels} for the robot pose. 
The plane can be parameterized as a 2 degree-of-freedom quaternion $\mathbf{q}^g_w$ and a scalar $d^g_w$ which represents the distance between the ground plane to the world frame origin. 
The residual is
\begin{equation}
\mathbf{r}_{\mathbf{p}} =  \left( \begin{array}{c}
\left(\mathbf{R}(\mathbf{q}^g_w) \mathbf{R}^w_i (\mathbf{R}^b_i)^\top \mathbf{e}_3 \right)_{1,2}\\
d^g_w + \mathbf{e}_3^\top \mathbf{R}(\mathbf{q}^g_w) (\mathbf{t}^w_i - \mathbf{R}^w_i {\mathbf{R}^b_i}^\top \mathbf{t}^b_i)
\end{array} \right),
\end{equation} 
with $\mathbf{e}_3 = \begin{pmatrix}0 & 0 & 1\end{pmatrix}^\top$. 
The plane motion error term becomes $E_{\mathit{plane}} = \sum_{l} \mathbf{r}_{plane,l}^\top \mathbf{\Sigma}_{plane,l}^{-1} \mathbf{r}_{plane,k}$ with covariance matrix $\mathbf{\Sigma}_{plane}$.
The stochasticity of the constraint allows for handling vibrations of the robot.

\subsection{Visual-Inertial Odometry with Motion Model Constraints}

We integrate the above introduced calibration parameters of the constraints as additional variables into the visual-inertial odometry.
The state of each frame in our optimization framework comprises the sensor pose $\mathbf{T}_{i}^w$, linear velocity $\mathbf{v}_t$, acceleration and gyroscope biases $\mathbf{b}_{acc}$, $\mathbf{b}_{gyro}$, landmark parameters $(u_l, v_l, d_l)$ of hosted keypoints, base frame to sensor frame extrinsics $\mathbf{T}_i^b$, and the global variables including the RBF parameters $\mu$, $\sigma$, $s$ and the plane parameters $\mathbf{q}^g_w$ and $d^g_w$. The optimized error function can be summarized as
$E = E_{\mathit{VIO}} + E_{\mathit{kin}} + E_{\mathit{plane}}$.
During optimization the extrinsic poses will be marginalized like linear velocity and IMU biases while the global variables are kept and their linearization point is fixed once the first connected state is marginalized. 
A discussion of the observability of our model is provided in appendix.

\section{EXPERIMENTS}

\begin{table*}[tbp!]
	\caption{Trajectory accuracy in RPE and ATE of our proposed approach (kin-vio) and a pure VIO method (vio).
	}
	\label{tab:result_tb_globaltoff}
	\begin{center}
		\begin{tabular}{ccccccccc}
			\toprule
			& \multicolumn{2}{c}{transl. RMSE RPE in\,m} &
			\multicolumn{2}{c}{rot. RMSE RPE in\,deg} &
			\multicolumn{2}{c}{transl. RMSE ATE in\,m} &
			\multicolumn{2}{c}{rot. RMSE ATE in\,deg}\\
			\cmidrule(lr){2-3} \cmidrule(lr){4-5} \cmidrule(lr){6-7} \cmidrule(lr){8-9}
			dataset & vio & kin-vio (ours) & vio & kin-vio (ours)  & vio & kin-vio (ours) & vio & kin-vio (ours)\\
			\midrule
			small-01 & 0.035 &  \textbf{0.021} & 0.659 & \textbf{0.597} & 0.037 & \textbf{0.014} & 0.713 & \textbf{0.463}\\
			small-02 & 0.120 & \textbf{0.106} & \textbf{0.664} & 0.756 & 0.097 & \textbf{0.077} & 0.656 & 
			\textbf{0.622}\\
			small-03 
			& 0.042 & \textbf{0.027} & 0.832 & \textbf{0.693} & 0.037 & \textbf{0.019} & 1.060 & \textbf{0.561}\\
			mid-01 & 0.232 & \textbf{0.197} & 1.439 & \textbf{1.242} & 0.190 & \textbf{0.153} & 0.978 & \textbf{0.957}\\
			mid-02
			& 0.195 & \textbf{0.158} & 1.171 & \textbf{1.100} & 0.150 & \textbf{0.108} & 0.807 & \textbf{0.739}\\
			mid-03 
			& 0.342 & \textbf{0.271} & 1.490 & \textbf{1.257} & 0.150 & \textbf{0.088} & 1.674 & \textbf{1.224}\\
			large-01 & 0.828 & \textbf{0.402} & 2.278 & \textbf{1.253} & 0.512 & \textbf{0.179} & 2.360 & \textbf{0.907}\\
			large-02 
			& {0.467} & \textbf{0.381} & 1.495 & \textbf{1.001} & {0.237} & \textbf{0.216} & 1.032 & \textbf{0.749}\\
			large-03 & 1.275 & \textbf{0.972} & 3.501 & \textbf{2.480}& 0.953 & \textbf{0.735} & 2.861 & \textbf{2.101}\\
			\bottomrule
		\end{tabular}
	\end{center}
\end{table*}

\begin{figure}[tb]
	\centering
	\includegraphics[width=0.4\linewidth]{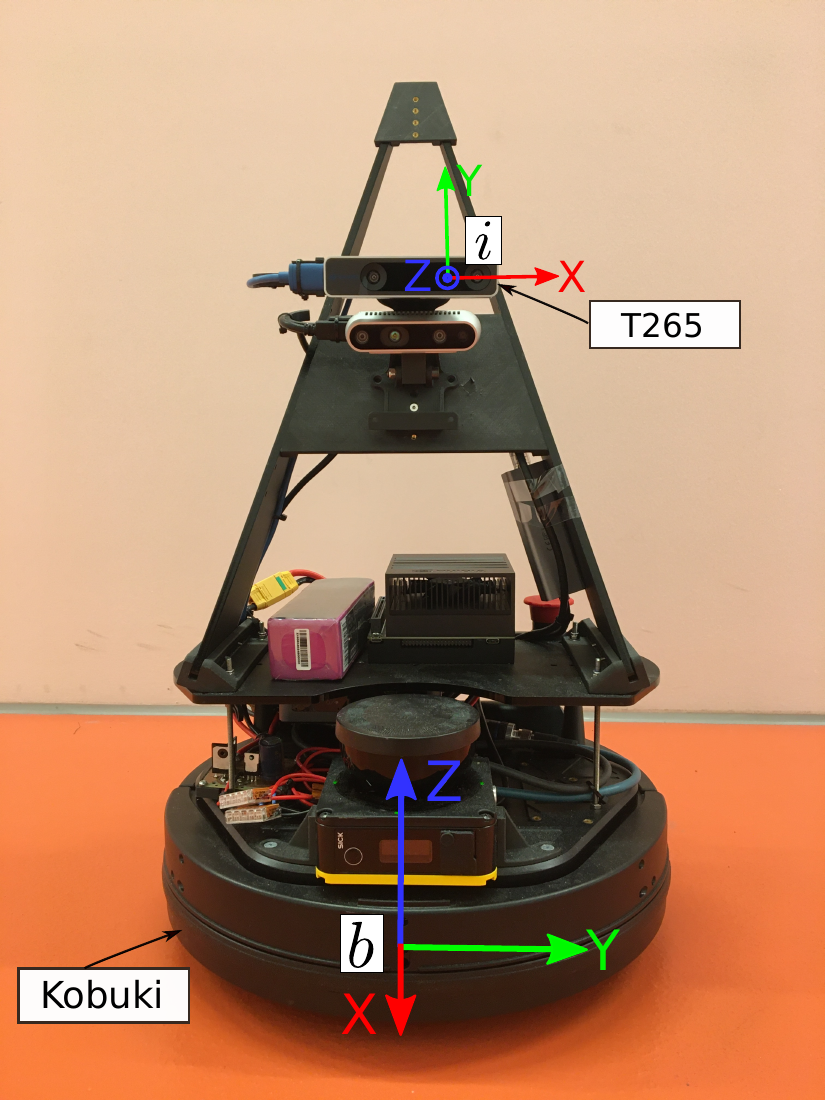}	
	\caption{Robot platform used in our experiments. The robot is built on a Kobuki mobile base with differential drive and is equipped with a Realsense T265 fisheye-stereo camera with IMU. The other sensor elements are not used in the experiment.}
	\label{fig:robot}
	
\end{figure}

We evaluate the proposed kinematics-constraint VIO on a differential drive robot with a fisheye-stereo camera and IMU (see Figs.~\ref{fig:robot}) in indoor environments. 
Similar as in~\cite{lee2020_viwo_onlinecalib} and~\cite{wu2017_vinsonwheels}, global offline optimization results are used as ground truth. 
To make sure enough loops can be found and the global optimization is accurate, the robot travels to the same location for several times in each recorded sequence.
We evaluate the accuracy of the estimate in terms of absolute trajectory (ATE), relative pose error (RPE)~\cite{Zhang18iros} and the error of the effective control velocity with the ground truth velocity. 
The RPE is computed by averaging the errors over $10,20,...,50\%$ sequence lengths of the full trajectory.
We also validate the prediction accuracy of the learned RBF kernel for different time horizons. 

Three groups of data with different environment scales are collected and each group consists of three different sequences. 
In the sequences, the robot starts from a static pose and then approximately drives at its maximum speed of 0.5 m/s. 
The robot traveled over wooden floor, concrete and tiles, which also cause vibrations on the robot. 
The average lengths are 57.2\,m (small scale), 222.3\,m (middle scale), and 413.3\,m (large scale). 
The ground truth for trajectory evaluation is computed using the global bundle adjustment layer of~\cite{usenko19nfr}. 
The method uses non-linear factor recovery to bound the computational and memory complexity of bundle adjustment using keyframes and to transfer information accumulated from intermediate frames and IMU measurements during VIO.
For calculating the effective control velocity error and prediction results, we use the dense global mapping result as ground truth where we set every frame as a keyframe and optimize them globally. 
The image rate is 30\,Hz, the IMU rate is 200\,Hz and the linear and angular commands are sent at the rate of 15\,Hz. 
The RBF parameters $\mu$, $\sigma$ and~$s$ are initialized to 0,~0.5 and 1 for both linear and angular velocity commands. A small command window can not collect enough information, while a large command window includes the commands that are far away from the current frame. We empirically choose a command window size of~3. 
The extrinsic parameters between base and sensor frame are initialized with values from the robot CAD model.

\subsection{Tracking Evaluation}

Table~\ref{tab:result_tb_globaltoff} summarizes the RPE and ATE evaluation results. 
By integrating the kinematics motion constraint and the plane constraint (kin-vio) both ATE and RPE are reduced over pure VIO (vio).  
Fig.~\ref{fig:rbf_traj} illustrates and compares the results of data sequence \emph{big\_01} estimated with purely VIO and our kinematics-constraint VIO. 
As can be seen, with the proposed method the deviation is decreased, especially in those parts of the trajectory with larger rotational motion.

In an ablation study, we compare the estimation accuracy of using RBF kernel weighting (kin-vio rbf) with other different weighting methods including RBF kernel with fixed initial parameters (kin-vio rbf w/o opt), non-weighted averaging of the command window (kin-vio avg), and using the last command that comes before the first frame of each frame pair (kin-vio raw). 
In addition, we also evaluate the estimate with only the plane motion constraint in addition to the VIO constraints. 
The error values are averaged over all data sequences and summarized in Tab.~\ref{tab:abl_tb1}. in which our combined method (kin-vio) consistently outperforms the others. The motion model can improve the tracking accuracy while the parameters are calibrated together with VIO. This can be attributed to the regularization by the parametric motion model whose parameters are adapted for a range of positive and negative velocity commands.

\begin{table*}[tbp!]
	\caption{Average trajectory accuracy in RPE and ATE and velocity error for different constraints over all sequences.}
	\label{tab:abl_tb1}
	\begin{center}
		\renewcommand{\tabcolsep}{5pt}
		\begin{tabular}{lcccccccccc}
			\toprule
			&\multicolumn{5}{c}{avg. transl. RMSE in\,m} &
			\multicolumn{5}{c}{avg. rot. RMSE in\,deg} \\
			\cmidrule(lr){2-6} \cmidrule(lr){7-11}
			&\makecell{kin-vio \\ ours: rbf \\ (w/o plane)} & \makecell{ kin-vio \\ rbf w/o opt \\ (w/o plane)}& \makecell{ kin-vio\\avg \\ (w/o plane)} & \makecell{ kin-vio\\raw \\ (w/o plane)} & \makecell{kin-vio \\ only plane} &
			\makecell{kin-vio \\ ours: rbf \\ (w/o plane)} & \makecell{ kin-vio \\ rbf w/o opt\\ (w/o plane)}& \makecell{ kin-vio\\avg \\ (w/o plane)} & \makecell{ kin-vio\\raw \\ (w/o plane)} & \makecell{kin-vio \\ only plane} \\
			\midrule
			RPE &\textbf{0.282} & 0.324 & 0.336 & 0.337 & 0.296 & \textbf{1.153} & 1.180 & 1.183 & 1.184 & 1.164\\
			& (\textbf{0.393}) & (0.441) & (0.452) & (0.452) &  & (\textbf{1.507}) & (1.592) & (1.600) & (1.601) &\\
			\midrule
			ATE & \textbf{0.177} & 0.210 & 0.219 & 0.220 & 0.190 & \textbf{0.925} & 0.947 &  0.948 & 0.950 & 0.935\\
			& (\textbf{0.270}) & (0.303) & (0.311) & (0.311) & & (\textbf{1.356}) & (1.413) & (1.418) & (1.418) &\\
			\midrule
			&\multicolumn{5}{c}{avg. RMSE of linear velocity in\,m/s} 
			&\multicolumn{5}{c}{avg. RMSE of angular velocity in\,deg/s} \\
			\cmidrule(lr){2-6} \cmidrule(lr){7-11}

			vel. error	&\textbf{0.025} & 0.034 & 0.036 & 0.036 & & \textbf{0.031} & 0.062 & 0.064 & 0.067 &\\
			\bottomrule
		\end{tabular}
	\end{center}
\end{table*}

\begin{figure*}[tb]
	\centering
	\subfloat[Prediction on dataset \emph{small-01}]{\includegraphics[width=.3\linewidth]{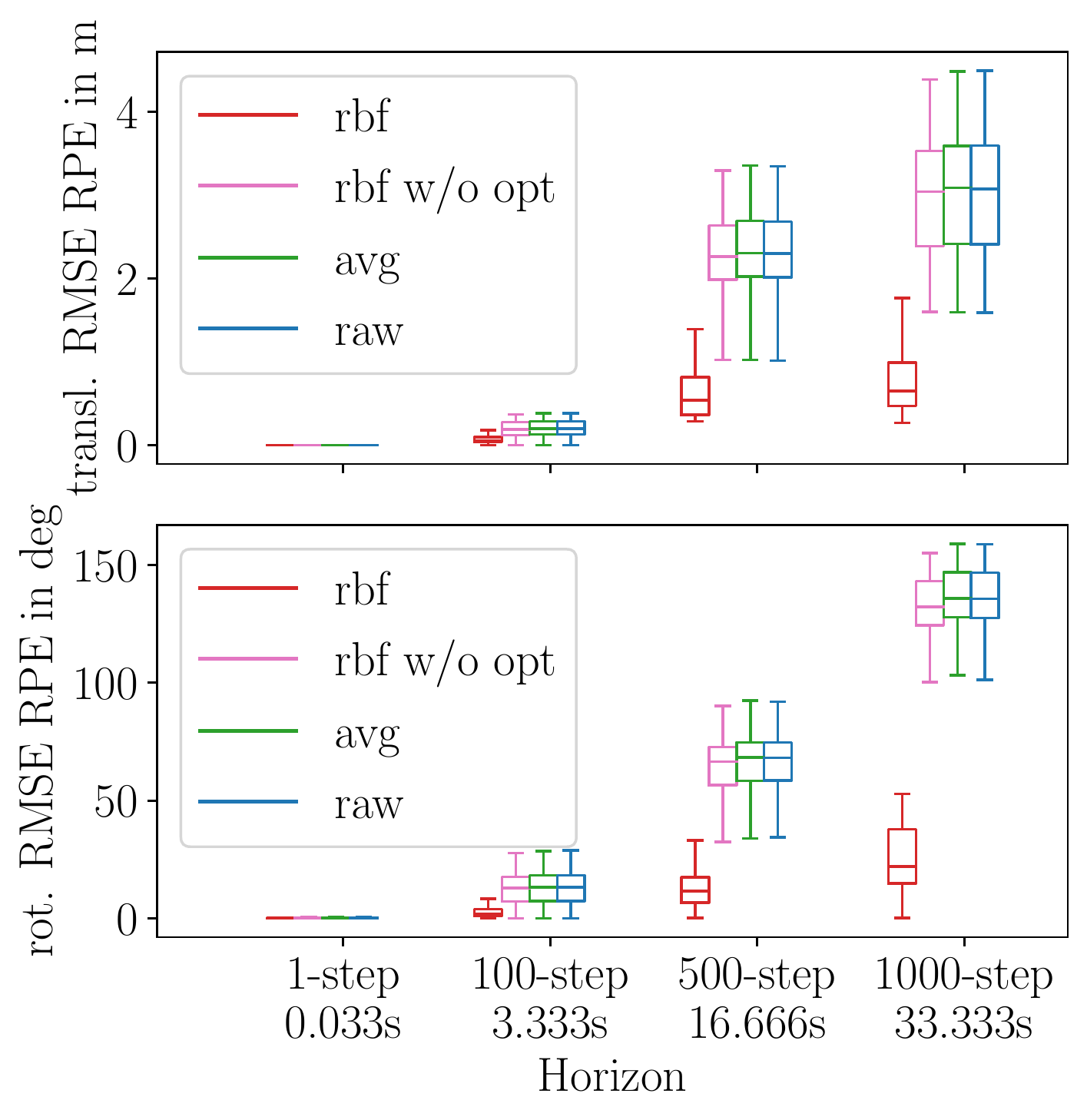}}  \label{fig:1a}
	\subfloat[Prediction on dataset \emph{mid-01}]{\includegraphics[width=.3\linewidth]{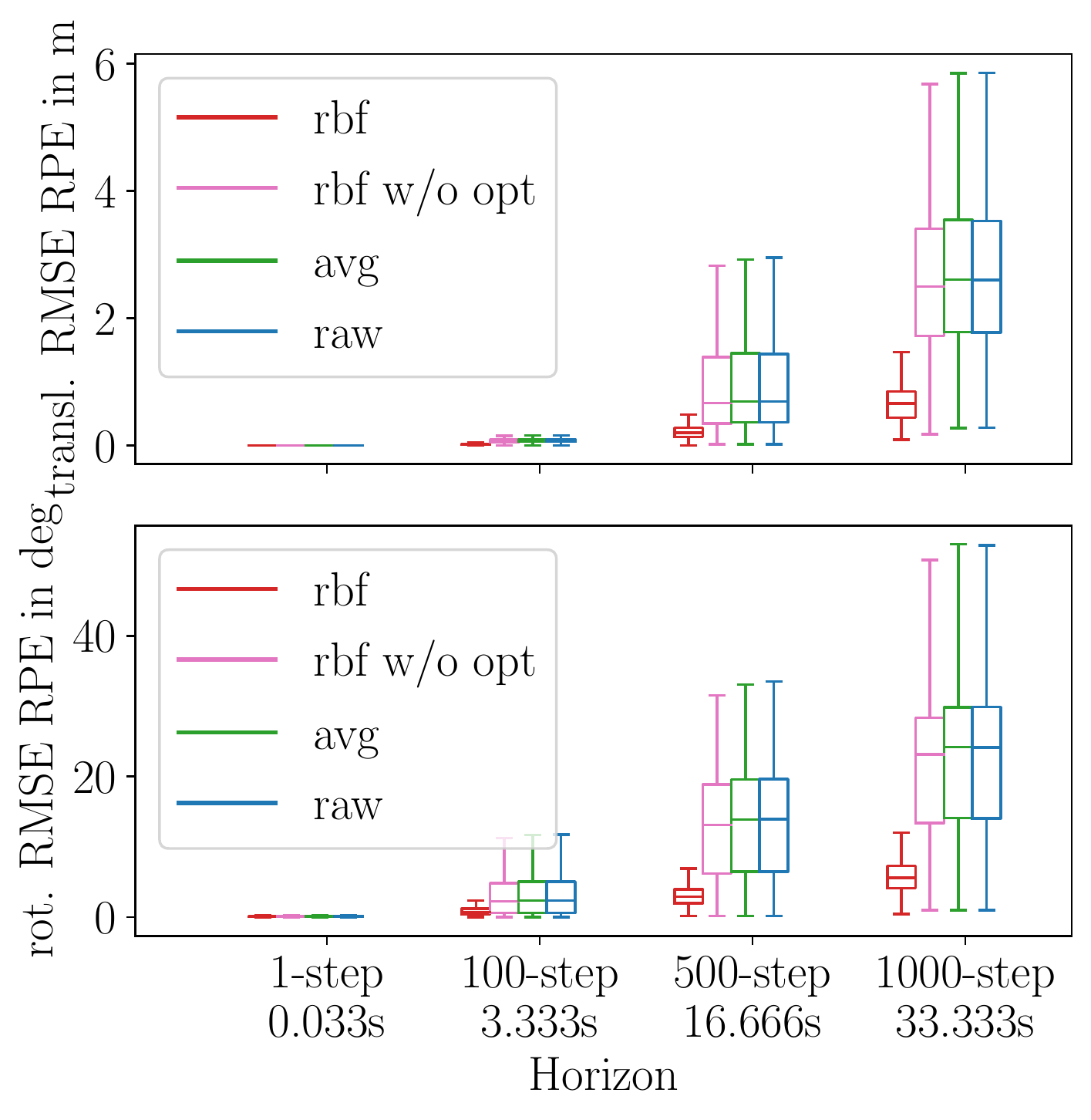}} \label{fig:1b}
	\subfloat[Prediction on dataset \emph{large-01}]{\includegraphics[width=.3\linewidth]{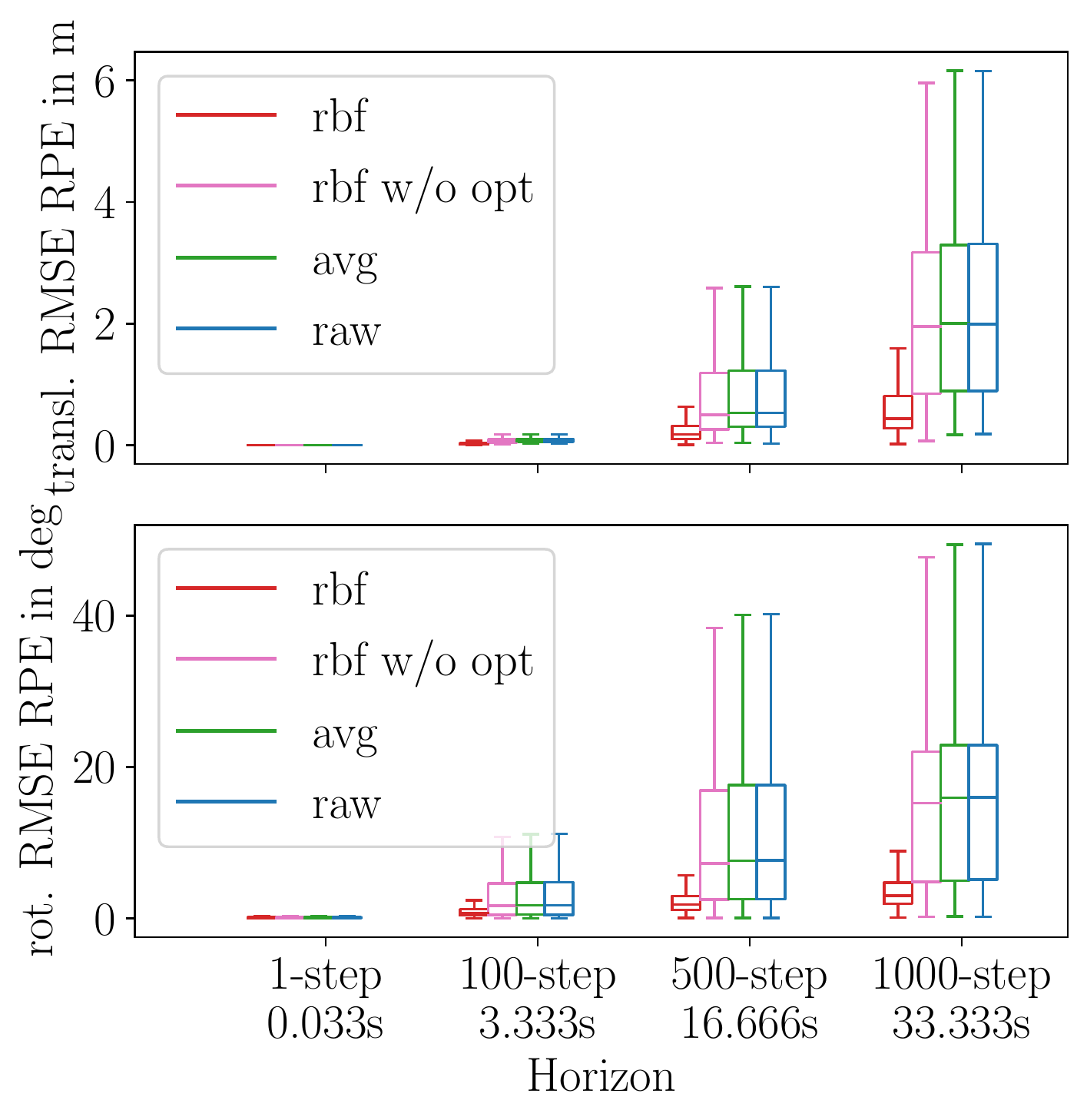}} \label{fig:1c}
	\caption{Prediction error on \emph{small-01}, \emph{mid-01} and \emph{large-01}. Our approach consistently has smallest prediction error.}
	\label{fig:pred_stats}
\end{figure*}

\begin{figure}[tb!]
	\centering
	\includegraphics[trim={0 0 0.05cm 0.05cm}, clip, width=0.89\linewidth]{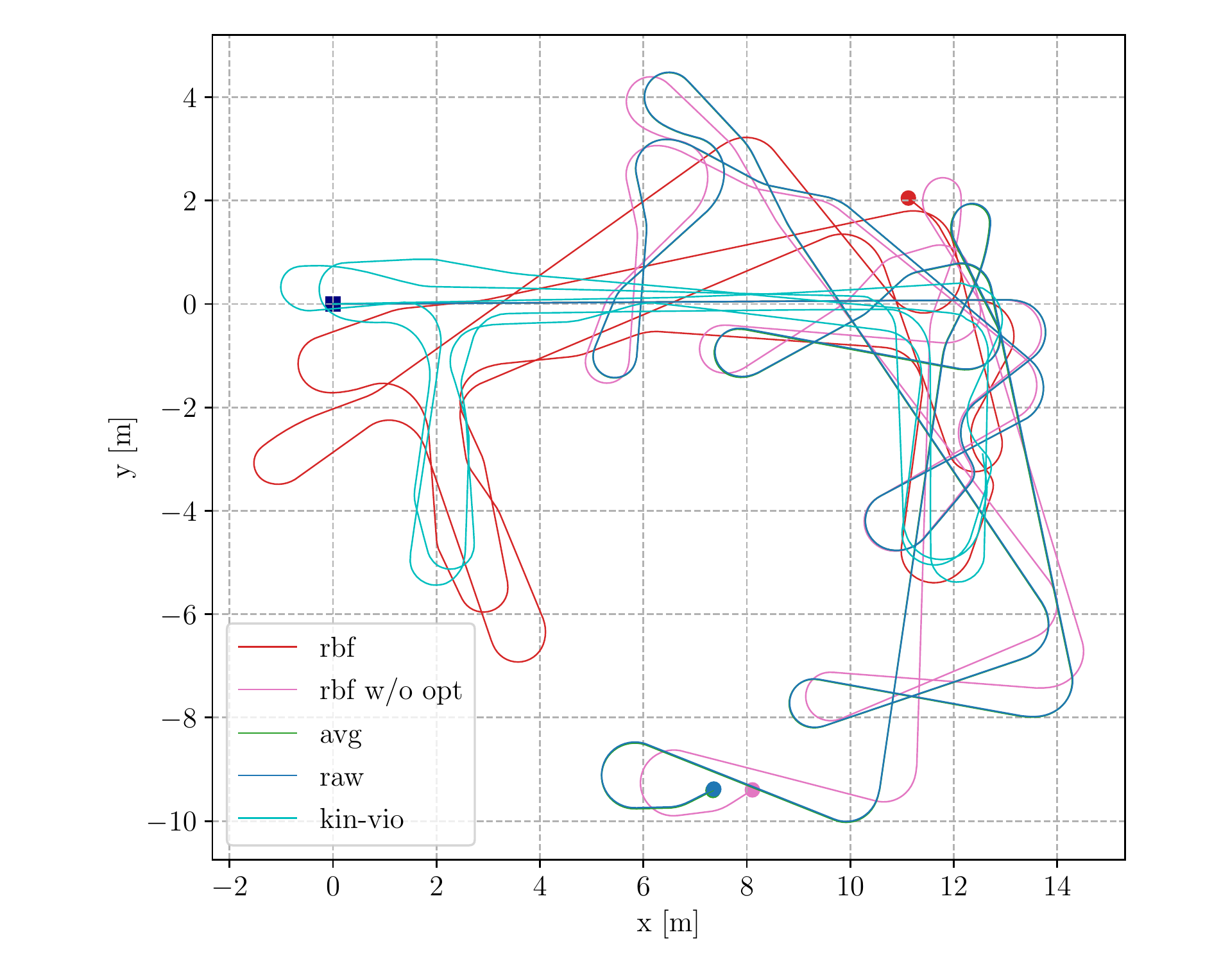}	
	\caption{Predicted trajectories from start (square) to end (circle) on \emph{small-02}. Our approach (rbf) follows the kinematics-constrained VIO result (kin-vio) closer.}
	\label{fig:pred_small_02}
\end{figure}

\subsection{Prediction Evaluation}

We also evaluate the performance of our method for forward motion prediction. 
Results for online prediction on data sequence \emph{small-01}, \emph{mid-01} and \emph{large-01} are shown in Fig.~\ref{fig:pred_stats}. 
For this evaluation, we compute the prediction from each frame in the trajectory with different horizon lengths starting from the current estimate of the parameters.
The command window is shifted along the future trajectory with the prediction step. 
We also compare the prediction accuracy with three alternative calibration methods. 
The first method uses the RBF model with constant initial parameters (rbf w/o opt). 
The second method uses the unweighted average value of the commands in the command window (avg). 
The third method calculates the prediction with the latest command (raw). 
Our proposed approach with calibrated RBF kernel parameters is denoted as "rbf".

It can be observed that the RBF kernel with optimized parameters has the smallest prediction error especially for longer prediction horizons. 
During the experiment we noticed that the improvement of the prediction accuracy is relatively smaller on the longer data sequences. 
This is because the rotations introduce larger errors and in the longer data sequences the robot mostly performs translational motion along the corridors with constant speed.   
Fig.~\ref{fig:pred_small_02} illustrates a prediction result on data sequence \emph{small-02} from beginning to the end of the trajectory with the final optimized RBF parameters. 
Our approach shows the smallest deviation compared with other methods.

\subsection{Computation Time}
We compare the run-time of our approach with the base VIO on our dataset using one Intel Xeon Silver 4112 CPU@2.60GHz with 8 cores.
On average the computation time increases by 34.14\% from 14.79~ms to 19.84~ms for processing one frame, the maximum time stays similar with 60.05~ms at rare peaks, the minimum time is 3.90~ms.  
The approach can still process faster than real-time.

\subsection{Discussion and Limitations}

A potential limitation of our approach can be estimation bias in the VIO in settings such as texture-less scenes or biased camera intrinsics. 
In our work, we assume that the VIO result is sufficiently accurate with negligible systematic offsets so it can be used to calibrate the effective control. 
In future work, we could additionally integrate other types of sensors like GPS. 
Outlier measurements can be handled with robust norms in the VIO.
One could also only activate the motion model when state variables like IMU bias are converged and indicate an accurate VIO. 

Integrating wheel encoder information can also improve VIO accuracy, as it measures the actual rotation of the wheels at high frequency. 
To convert the wheel encoder information to body velocity or relative pose, one needs to consider the type of the vehicle.
Our approach is simpler to integrate for robots whose motion can be modeled with our model (such as differential or Ackerman drives \cite{thrun2005probabilistic}). 
Moreover, our calibrated model could be used for downstream tasks such as model-predictive control based path tracking.
In future work, we are also interested in combining wheel encoder measurements with our method.

\section{CONCLUSIONS}
In this paper we present a VIO approach based on non-linear windowed optimization that includes velocity-control based kinematic motion model constraints.
The motion model is integrated as a new factor between each image pair. 
We compute the 2D robot velocity between two consecutive images from the state estimate and compare it against the control command sent to the robot. 
To compensate the difference between the control command and the real robot action, we use RBF kernels along the time domain to determine an effective control command from the raw commands and calibrate the RBF parameters online in the VIO system. 
Our experiments demonstrate that by using this motion constraint in addition to a planar motion constraint not only the accuracy of the VIO is improved but the learned motion model can also predict the robot motion more accurately. In future work, more complex motion models or including other sensors like GPS or wheel odometry could be investigated.

\section*{APPENDIX}
\subsection{Inverse vs. Forward Model Residuals}
\label{app:forwinvmodel}

Given the relative pose between two frames $
\mathbf{T}_{b,t'}^{b,t}$, the corresponding $SE(2)$ pose $\mathbf{P}_{b,t'}^{b,t}$ in the horizontal plane consisting of rotational part $\mathbf{Q}_{b,t'}^{b,t}$ and translational part $\mathbf{p}_{b,t'}^{b,t}$,
the effective control input $\boldsymbol{\xi} = (v, 0, w)^\top$ with linear and angular velocity, and $\bar{\mathbf{P}}^{b,t}_{b,t'} = exp(\Delta t \boldsymbol{\xi})$, the forward model can be written as
\begin{equation}
\mathbf{r}_{\mathit{forw}} = f({\mathbf{P}_{b,t'}^{b,t}} , \bar{\mathbf{P}}^{b,t}_{b,t'}) =
\begin{bmatrix}
\mathbf{p}_{b,t'}^{b,t} - \bar{\mathbf{p}}_{b,t'}^{b,t}\\
log_{\mathit{so2}}(\mathbf{Q}_{b,t'}^{b,t}) - w \Delta t
\end{bmatrix},
\end{equation}
with derivatives $\frac{\partial \mathbf{r}_{\mathit{forw}}}{\partial \mathbf{P}_{b,t'}^{b,t}} = \frac{\partial f(.)}{\partial {\mathbf{P}_{b,t'}^{b,t}}}$ 
and $
\frac{\partial \mathbf{r}_{\mathit{forw}}}{\partial \mathbf{p}_{\mathit{rbf}}} = \frac{\partial f(.)}{\partial \bar{\mathbf{P}}^{b,t}_{b,t'}} \frac{\partial \bar{\mathbf{P}}^{b,t}_{b,t'}}{\partial \boldsymbol{\xi}}
\frac{\partial \boldsymbol{\xi}}{\partial \mathbf{p}_{\mathit{rbf}}}$, while the inverse model is
\begin{equation}
\mathbf{r}_{\mathit{inv}} = log_{\mathit{se2}}({\mathbf{P}_{b,t'}^{b,t}}) - \Delta t \boldsymbol{\xi}
= \begin{bmatrix}
log_{\mathit{se2}}({\mathbf{P}_{b,t'}^{b,t}})_{x,y} - v \Delta t\\
log_{\mathit{so2}}(\mathbf{Q}_{b,t'}^{b,t}) - w \Delta t
\end{bmatrix},
\end{equation}
with derivatives $\frac{\partial \mathbf{r}_{\mathit{inv}}}{\partial \mathbf{P}_{b,t'}^{b,t}} = \frac{\partial log_{\mathit{se2}}(.)}{\partial \mathbf{P}_{b,t'}^{b,t}}$ and $
\frac{\partial \mathbf{r}_{\mathit{inv}}}{\partial \mathbf{p}_{\mathit{rbf}}} = -\frac{\partial \boldsymbol{\xi}}{\partial \mathbf{p}_{\mathit{rbf}}} \Delta t$.

\subsection{Observability}
\label{app:observability}

As in~\cite{wu2017_vinsonwheels}, the observability of the state variables can be analyzed based on the underlying state-space model irrespective of the implementation of the estimator. 
We discuss the observability properties of the state variables such as pose, plane parameters and RBF parameters for our model. A detailed analysis can be found in~\cite{obsReport}.
We follow the derivation in~\cite{wu2017_vinsonwheels}.
In general, the observability properties of the pose variables are the same for forward and inverse model. 
While in the forward model the derivative wrt. $\mathbf{p}_{b,t'}^{b,t}$ is an identity matrix, the derivative in the inverse model is $J_{\log_{\mathit{se2}}}$.
Because $J_{\log_{\mathit{se2}}}$ is an invertible matrix by definition,
when we compute the observability matrix by multiplying this Jacobian matrix and the transition matrix, 
the rank of the observability matrix remains the same based on Sylvester’s inequality. 
Note that we use a stereo camera and hence scale becomes directly observable.
We follow the proof scheme in~\cite{wu2017_vinsonwheels} and show that the global orientation becomes observable by using the plane constraint and a set of priors in the initial frames. 
As in~\cite{wu2017_vinsonwheels}, the state transition is given by the IMU propagation model. 
The constraint from the velocity-based kinematic motion model is treated as an observation.
The linearized transition matrix $\mathbf{\Phi}_{k,1}$  from time-step 1 to $k$  for the IMU propagation is derived in~\cite{hesch2014_vio_analysis}.
By including the transition of the RBF parameters as a constant propagation model, the transition matrix of our model becomes:
$\bar{\boldsymbol{\Phi}}_{k,1} = \begin{bmatrix}
\mathbf{\Phi}_{k,1} & \mathbf{0}\\
\mathbf{0} & \mathbf{I}
\end{bmatrix}$
with the augmented state vector including plane parameters $\mathbf{q}^{g}_{w}$, $d^{g}_{w}$ between ground $g$ and world frame $w$, extrinsic parameters $\mathbf{q}^{b}_i$, $\mathbf{t}^{b}_i$ between robot base and IMU frame and RBF parameters $\mathbf{p}_{\mathit{rbf}}$, 
$\bar{\mathbf{x}} = \begin{bmatrix}
\mathbf{x}^{\top} & \mathbf{q}^{g}_{w} &d^{g}_{w} & \mathbf{q}^{b}_i & \mathbf{t}^{b}_i & \mathbf{p}_{\mathit{rbf}}
\end{bmatrix}^{\top}$.

As shown in~\cite{wu2017_vinsonwheels}, the plane distance $d^{g}_{w}$ is observable. 
Following the derivation in~\cite{wu2017_vinsonwheels}, the global orientation of the plane can be shown to be unobservable. 
Since our robot starts from still state, we use the initial accelerator measurement to initialize the plane angle. 
By adding this information as a Gaussian initial prior on the plane angle $\mathbf{q}^{g}_{w}$, the global orientation becomes observable. 
As shown in~\cite{lee2020_viwo_onlinecalib} the extrinsic parameters are also unobservable under specific motions. 
We used the extrinsics derived from the CAD model as an initial Gaussian prior to counteract this problem. 
We prove the observability of the RBF parameters ($s_{lin}, \mu_{lin}, \sigma_{lin}$) for the translational velocity. 
The parameters for the angular velocity have the same observability. We denote $\exp\left( - \frac{\| d\tau_i - \mu_{lin} \|^2}{2 \sigma_{lin} ^2} \right)$ as $\exp(.) $, where $d\tau_i$ is the time difference between control command at $t_i$ and image frame at $t$.
The Jacobian matrix is $\mathbf{H}_{rbf} = 
\begin{bmatrix}
\dots & -\frac{\bar{v}}{s} & A & B
\end{bmatrix}$,
\begin{align}
\bar{v} &:= 	s \frac{\sum_{i = 1}^{N} \exp(.) v_{t_i}} {\sum_{i = 1}^{N} \exp(.)},\\
A &:= \frac{\bar{v} {\sum_{i = 1}^{N} \exp(.)} \frac{d\tau_i - \mu}{\sigma^2} - s  {\sum_{i = 1}^{N} v_i \exp(.)} \frac{d\tau_i - \mu}{\sigma^2}}{{\sum_{i = 1}^{N} \exp(.)}},\\
B &:= \frac{\bar{v} {\sum_{i = 1}^{N} \exp(.)} \frac{(d\tau_i - \mu)^2}{\sigma^3} - s  {\sum_{i = 1}^{N} v_i \exp(.)} \frac{(d\tau_i - \mu)^2}{\sigma^3}}{{\sum_{i = 1}^{N} \exp(.)}}.
\end{align}
The nullspace for the RBF parameters $\mathbf{N}_{\mathit{rbf}}$ is 
$\mathbf{N}_{rbf} = \begin{bmatrix}
\mathbf{0} & \dots & \mathbf{0} &\mathbf{I}_{3\times 3}
\end{bmatrix}^\top$. The corresponding $3 \times 3$ bottom right block of the observability matrix for our inverse motion model is
$\mathbf{M}_{\mathit{rbf}} = \begin{bmatrix}
-\frac{\bar{v}}{s} & 0 & 0 \\
0 & A& 0 \\
0 & 0 & B
\end{bmatrix}$.
The product of the motion model observability matrix and the nullspace $\mathbf{N}_{\mathit{rbf}}$ is equal to $\mathbf{M}_{\mathit{rbf}}$ as the other terms cancel out by multiplication with the $\mathbf{0}$ components in $\mathbf{N}_{\mathit{rbf}}$.
The RBF parameters are unobservable if the velocities $v_i$ in the window are constant, which can happen especially at the beginning of the datasets when the robot stands still.
We placed a weak Gaussian prior on the initial values of the RBF parameter estimates.  
Due to the marginalization prior, the RBF parameters will remain observable even if they become temporarily unobservable in the window.

\end{document}